\documentclass[letterpaper, 10 pt, conference]{ieeeconf}  %

\IEEEoverridecommandlockouts                              %

\usepackage[dvipsnames]{xcolor}%
\usepackage{subcaption}
\usepackage[colorinlistoftodos,prependcaption,textsize=footnotesize]{todonotes}
\presetkeys{todonotes}{inline}{}   %

\usepackage[acronym]{glossaries}
\glsdisablehyper   %
\newacronym{3dssg}{3DSSG}{3D Semantic Scene Graph}
\newacronym{clip}{CLIP}{Contrastive Language-Image Pretraining}
\newacronym{gnn}{GNN}{Graph Neural Network}
\newacronym{xai}{XAI}{Explainable AI}
\newacronym{tsdf}{TSDF}{Truncated Signed Distance Function}
\newacronym{vlfm}{VLFM}{Vision-Language Foundation Model}
\newacronym{llm}{LLM}{Large Language Model}
\newacronym{dsg}{DSG}{3D Dynamic Scene Graphs}
\newacronym{hm3d}{HM3DSEM}{Habitat-Matterport 3D Semantics Dataset} %

\usepackage{xspace}
\newcommand{\eg}{e.\,g.,\xspace}

\usepackage{tabularx}
\usepackage{booktabs}
\usepackage{multirow}
\usepackage{siunitx}
\usepackage{amssymb}
\usepackage{url}
\usepackage{censor}
\let\labelindent\relax
\usepackage{enumitem}

\StopCensoring

\title{\LARGE \bf
DISC: Dense Integrated Semantic Context for Large-Scale Open-Set Semantic Mapping
}

\author{\blackout{Felix Igelbrink$^{1}$ and Lennart Niecksch$^{2, 1}$ and Martin Atzmueller$^{2, 1}$ and Joachim Hertzberg$^{2, 1}$%
\thanks{$^{1}$Felix Igelbrink, Lennart Niecksch and Martin Atzmueller are with the German Research Center for Artificial Intelligence (DFKI), Osnabrück, Germany
        {\tt\small <firstname>.<lastname>@dfki.de}}%
\thanks{$^{2}$Lennart Niecksch, Martin Atzmueller and Joachim Hertzberg are with the Department of Computer Sciene, Osnabrück University,
        Osnabrück, Germany
        {\tt\small <firstname>.<lastname>@uni-osnabrueck.de}}}%
}

\begin{document}

\maketitle
\thispagestyle{empty}
\pagestyle{empty}

\begin{abstract}
Open-set semantic mapping enables language-driven robotic perception, but current instance-centric approaches are bottlenecked by context-depriving and computationally expensive crop-based feature extraction. To overcome this fundamental limitation, we introduce DISC (Dense Integrated Semantic Context), featuring a novel single-pass, distance-weighted extraction mechanism. By deriving high-fidelity CLIP embeddings directly from the vision transformer's intermediate layers, our approach eliminates the latency and domain-shift artifacts of traditional image cropping, yielding pure, mask-aligned semantic representations. To fully leverage these features in large-scale continuous mapping, DISC is built upon a fully GPU-accelerated architecture that replaces periodic offline processing with precise, on-the-fly voxel-level instance refinement. We evaluate our approach on standard benchmarks (Replica, ScanNet) and a newly generated large-scale-mapping dataset based on Habitat-Matterport 3D (\acrshort{hm3d}) to assess scalability across complex scenes in multi-story buildings. Extensive evaluations demonstrate that DISC significantly surpasses current state-of-the-art zero-shot methods in both semantic accuracy and query retrieval, providing a robust, real-time capable framework for robotic deployment. The full source code, data generation and evaluation pipelines will be made available at \censor{\url{https://github.com/DFKI-NI/DISC}}.
\end{abstract}

\section{INTRODUCTION}

Semantic mapping has long been a foundational component for autonomous mobile robots, enabling them to understand and interact with complex environments \cite{nuchterSemanticMapsMobile2008b,kostavelis2015semantic}.
While classical approaches relied on rigid, predefined, closed-set vocabularies, the recent integration of \glspl{vlfm}, most notably CLIP \cite{radford2021learning}, has boosted the paradigm towards open-set embedding spaces \cite{igelbrink2024online}. 
By anchoring these open-vocabulary features into 3D spatial representations, robots are enabled to perform complex, language-driven tasks, ground textual concepts into geometry, and answer free-form queries.

Early integrations of VLFMs primarily utilized implicit representations (\eg NeRFs) or dense point clouds. While these approaches allow for natural language querying on a geometric basis, they often lack an explicit object-level abstraction. Consequently, instance-based, object-centric mapping and \glspl{3dssg} have emerged as the preferred representation for high-level reasoning. However, for utilization on an actual mobile agent instead of being restricted to offline processing, this mapping process must be incremental, fast, and capable of scaling to large environments during continuous operation.
\begin{figure}[t]
  \centering
  \begin{subfigure}[c]{\linewidth}
    \centering
    \includegraphics[width=\textwidth]{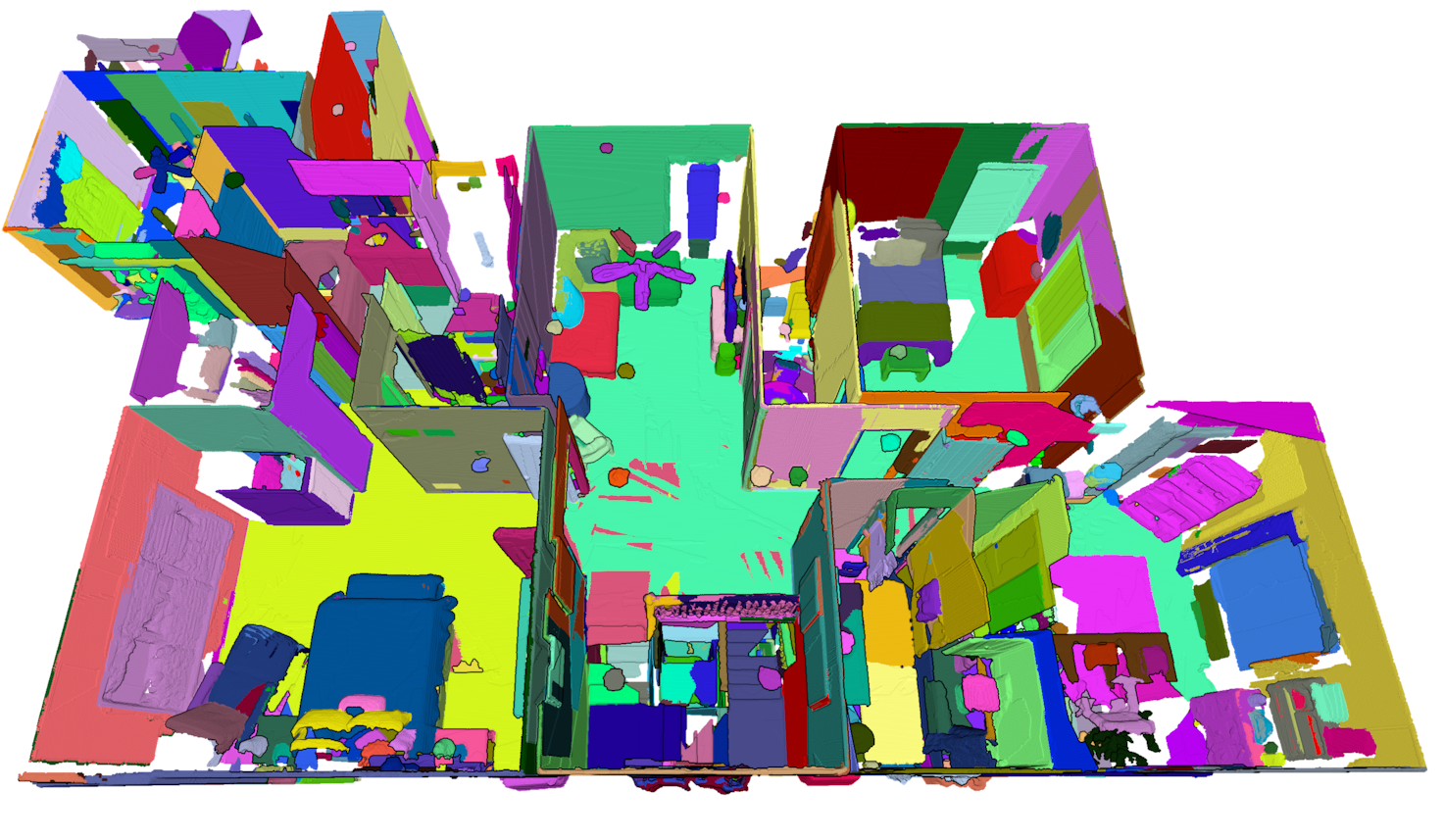}
  \end{subfigure}
  \begin{subfigure}[c]{\linewidth}
    \centering
    \includegraphics[width=\textwidth]{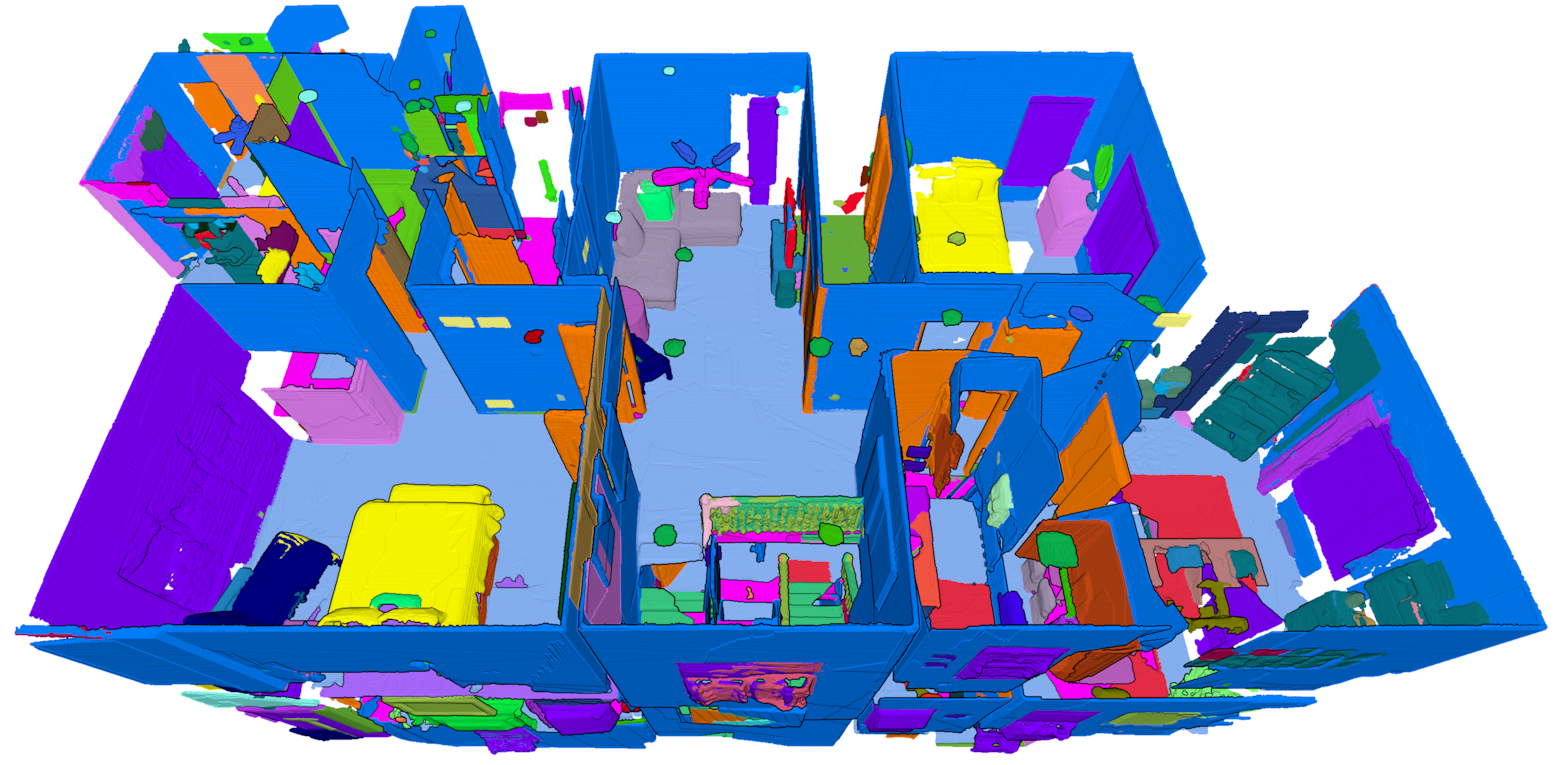}
  \end{subfigure}
  \begin{subfigure}[c]{\linewidth}
    \centering
    \includegraphics[width=\textwidth]{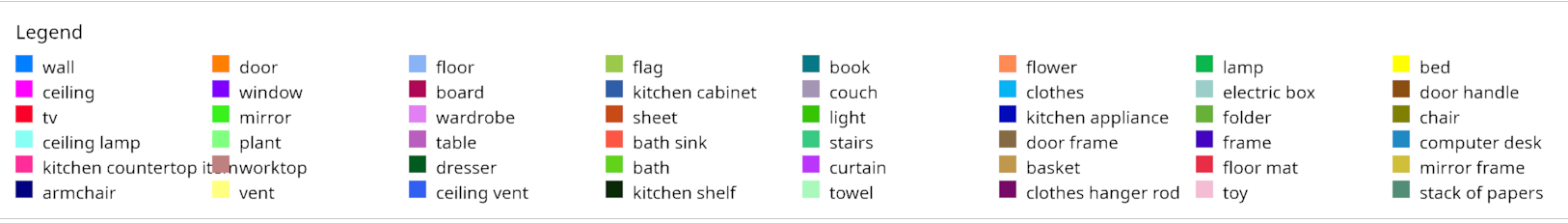}
  \end{subfigure}
  \caption{Example mapping results on a \acrshort{hm3d} scene. Top: The tracked instances of the map, randomly colored. Bottom: The resulting semantic segmentation with the $\text{top}_k$ semantic classes.
  \label{fig:map_results}}
\end{figure}
Despite significant recent advancements, deploying object-centric open-set systems at scale remains challenging.
Most existing works are constrained to small, single-room environments due to the high computational overhead associated with map generation and feature extraction. 
Recent state-of-the-art approaches~\cite{werby2024hierarchical,Gu2024-bq,linok2025beyond,mirzaei2025core} have achieved impressive results, but typically encounter two primary bottlenecks. 
First, to maintain sufficient performance, they rely on fast, but imprecise heuristics, e.g., bounding-box-based overlaps for data association and require periodic, com\-pu\-ta\-tion\-al\-ly expensive, offline refinement stages to resolve over-segmentation artifacts. 
Second, and more importantly, their reliance on crop-based image extraction for obtaining instance-level CLIP features introduces fundamental representational flaws. Foundation models such as CLIP are pre-trained on complete, natural images. When these are aggressively cropped or their backgrounds are artificially masked out, the resulting inputs deviate significantly from the model's training distribution. This severe domain shift actively degrades the model's zero-shot classification capabilities \cite{liang2023open}. Furthermore, the crop-and-recompute pipeline removes the global image context, which is crucial for CLIP to correctly resolve semantic ambiguities and spatial relationships~\cite{Yu_2023_CVPR}.

To address these limitations in semantic mapping, we introduce DISC (Dense Integrated Semantic Context), a fully GPU-accelerated mapping architecture designed for large-scale, open-set semantic mapping. 
DISC eliminates the need for periodic offline refinement by directly utilizing accurate voxel-level overlap metrics for immediate, per-frame local refinement. This allows the system to efficiently merge instances on-the-fly when sufficient geometric evidence is present. Furthermore, we present a %
single-pass feature extraction mechanism inspired by MaskCLIP \cite{dong2023maskclip}. We extract dense patch-level features directly from the intermediate transformer layers of the CLIP model, circumventing the need for costly and noise-prone instance cropping entirely. Finally, to rigorously evaluate open-set semantic mapping across extensive, multi-story indoor environments, we introduce a new benchmark dataset derived from \acrfull{hm3d} \cite{ramakrishnan2021habitat}. Figure~\ref{fig:map_results} shows an example mapping result on this dataset.

In summary, our main contributions are:
\begin{itemize}
\item A fully GPU-accelerated 3D semantic mapping pipeline that leverages direct voxel-overlap for fast, incremental, and continuous instance refinement in large-scale environments.
\item A method to derive and integrate high-fidelity CLIP features directly from the model without cropping. This utilizes an incremental, geometry-based quality fusion mechanism to ground open-vocabulary queries efficiently.
\item A new HM3D-based evaluation protocol and dataset to benchmark semantic mapping in large, multi-room indoor environments, alongside competitive performance on established benchmarks.
\end{itemize}

\section{RELATED WORK}

\subsection{Open-Set Semantic Mapping}

Recent advancements in open-set semantic mapping can be broadly categorized into implicit, geometry-dense representations and explicit, object-centric approaches.

Early approaches focused on integrating open-vocabulary features directly into continuous or dense map representations. For instance, LERF \cite{kerrLERFLanguageEmbedded2023} embeds language features into Neural Radiance Fields (NeRFs), while approaches like OpenScene \cite{pengOpenScene3DScene2023a} and ConceptFusion \cite{jatavallabhula2023conceptfusion} project dense pixel-level multi-modal features onto 3D point clouds and meshes. While these methods successfully enable natural language querying, they treat the environment as a monolithic volume or a raw collection of points. The resulting lack of an explicit object-level abstraction limits downstream reasoning tasks, such as task planning or dynamic scene graph generation.

To bridge this gap, more recent works have focused on instance-based, object-centric mapping. Approaches like ConceptGraphs \cite{Gu2024-bq}, Beyond Bare Queries (BBQ) \cite{linok2025beyond}, Core3D \cite{mirzaei2025core}, and HOV-SG \cite{werby2024hierarchical} construct explicit object maps and 3D Semantic Scene Graphs (3DSSGs) that support complex queries and task-dependent clustering natively~\cite{maggio2024clio}. 
Our work builds upon this object-centric philosophy. However, while existing methods rely heavily on CPU-bound spatial heuristics and offline processing steps, limiting their applicability, we fundamentally redesign the underlying architecture. By shifting the entire pipeline to the GPU, DISC is capable of mapping prodigious multi-story buildings incrementally and efficiently, resolving spatial associations on-the-fly, overcoming the computational bottlenecks of current state-of-the-art pipelines.

\subsection{Scalability and Instance Refinement}

A major challenge in instance-based mapping is the robust association of 2D object-agnostic segmentation masks obtained from \emph{Segment Anything} Models (SAM)~\cite{kirillov2023segment} (\eg FastSAM \cite{zhao2023fast} or MobileSAM \cite{zhang2023faster}) to 3D object nodes across frames. To maintain acceptable frame rates, many state-of-the-art systems rely on fast but coarse spatial heuristics, such as Axis-Aligned Bounding Box (AABB) overlaps, for greedy data association \cite{Gu2024-bq, linok2025beyond}. Because this fast matching inevitably leads to over-segmentation and temporal inconsistencies, these systems typically require a computationally expensive, periodic offline refinement stage to resolve conflicts and merge fragments.

In contrast, our architecture aims to break the reliance on this offline post-processing to obtain a usable map. Instead of using coarse AABB-metrics. our system computes precise voxel-based intersections utilizing GPU-based sparse matrix algorithms. This enables a highly efficient, per-frame refinement stage that merges instances in active regions on-the-fly as soon as sufficient geometric evidence is accumulated, drastically improving map consistency during online exploration without sacrificing mapping speed.

\subsection{Vision-Language Feature Integration}

Extracting instance-level semantic embeddings from \glspl{vlfm} is a major computational bottleneck. The prevailing strategy in open-set mapping involves cropping the RGB image using the 2D bounding boxes or instance masks, and then passing each crop independently through the CLIP encoder \cite{Gu2024-bq, jatavallabhula2023conceptfusion, Schmid-RSS-24}. 
Recently, the approach in \cite{mirzaei2025core} extended this to a coarse-to-fine mask generation approach, increasing the number of required CLIP inferences even more. 
While generally working well, the approach is not only prohibitively sumptuous for real-time mapping, especially with dozens of instances per frame, but it also discards the global image context, focusing only on a single, potentially small crop of the original frame. 
Additionally, as these crops often contain pixels from surrounding instances, the resulting feature vectors tend to contain information about nearby objects as well, \eg a \emph{wall} with a \emph{picture} on it usually also has a high similarity with a query for just \emph{picture}. 
This additionally limits the accuracy when using the CLIP features to associate new segments with existing ones as well~\cite{Yu_2023_CVPR}.

To circumvent this, recent works like BBQ \cite{linok2025beyond} utilize DINOv2 \cite{oquab2023dinov2} patch features for robust instance tracking but still rely on computationally heavy offline feature extraction for the final semantic CLIP representations. 
Inspired by this work as well as the MaskCLIP \cite{dong2023maskclip} approach, we aim at directly extracting dense patch-level features from the intermediate transformer layers on a single CLIP forward pass. 

\section{METHOD}

Our proposed architecture, DISC (Dense Integrated Semantic Context), establishes a fully GPU-accelerated \acrfull{3dssg} as the foundational representation for the mapping life-cycle. Unlike conventional pipelines that separate fast spatial tracking from computationally heavy offline semantic refinement, DISC is designed to be strictly incremental. It integrates high-fidelity visual features and precise geometric associations immediately during the online mapping process, ensuring the scene representation remains topologically and semantically valid at every time step. The backend data structure is an instance-based, object-centric map implemented as a multi-layered graph. By following a GPU-first principle and keeping the entire pipeline resident on the GPU, we eliminate the need for simplified bounding-box heuristics, enabling dense, voxel-level operations for every incoming frame.

\subsection{Segmentation and Feature Extraction}

For each incoming RGB-D frame, we extract a set of 2D instance segmentation masks. While our pipeline is fundamentally agnostic to the specific Segment Anything Model (SAM) architecture, we employ FastSAM \cite{zhao2023fast} for our experiments to achieve an optimal trade-off between segmentation quality and real-time inference speed. To filter out noise, masks with low confidence, extreme aspect ratios, or insufficient size are immediately discarded. 

Simultaneously, following \cite{linok2025beyond}, we extract dense visual feature vectors from the RGB image using the DINOv2 model \cite{oquab2023dinov2}. Because DINOv2 strongly emphasizes local visual structure, it provides robust features for instance tracking. By extracting patch-level features from the intermediate layers, we only need to execute the heavy feature encoder once per frame. The depth data corresponding to each valid segment is then projected into a 3D point cloud, filtered using a custom, parallelized CUDA implementation of the DBSCAN algorithm \cite{Macklin2022Warp} to remove noise, and subsequently voxelized.

\subsection{Data Association and Scene Integration}

The core advantage of our GPU-accelerated pipeline lies in how local segments are integrated into the global map. Traditional approaches rely on a tandem of fast, greedy AABB matching followed by periodic, expensive offline refinement passes over the entire map to fix over-segmentation. We replace this paradigm with a single per-frame active refinement stage based directly on 3D voxel overlap. 

When a new frame is processed, locally detected segments are matched against the global scene. We employ an efficient broad-phase Bounding Volume Hierarchy (BVH) collision detection to identify an \emph{active set} of candidate instances—existing map segments that potentially overlap with newly detected instances or their immediate neighborhood. Subsequently, we compute the exact geometric voxel intersection among all candidates. This ensures that at each refinement step, newly detected instances and nearby existing instances are merged if sufficient geometric evidence (voxel overlap) and visual similarity (cosine similarity) are met. By relying on robust DINO features to govern these structural merges, we prevent the catastrophic fusion of semantically distinct but spatially close objects. 

At the very end of the mapping trajectory, a final, lightweight post-processing step iterates over the map to merge remaining orphaned candidates and filter out residual noisy instances, such as segments containing fewer than a minimum threshold of voxels.

\subsection{Vision-Language Feature Integration}

\begin{figure}[tb]
  \centering
  \begin{subfigure}[c]{0.49\linewidth}
    \centering
    \includegraphics[width=\textwidth]{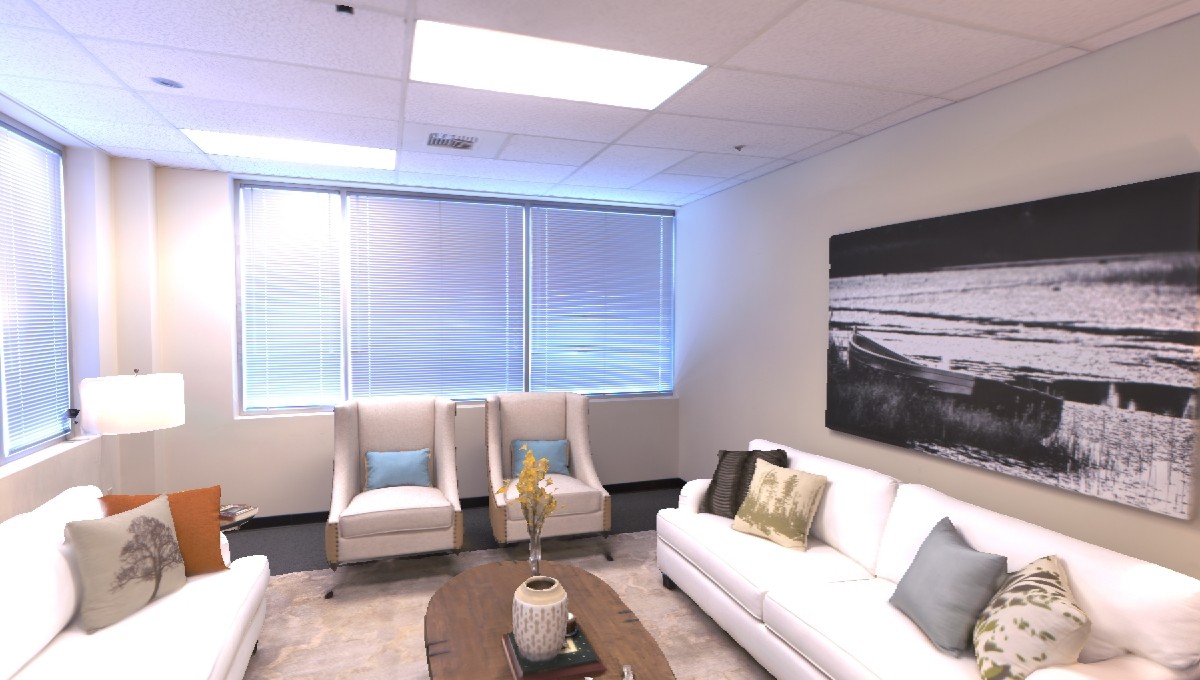}
    \subcaption[]{}
  \end{subfigure}
  \begin{subfigure}[c]{0.49\linewidth}
    \centering
    \includegraphics[width=\textwidth]{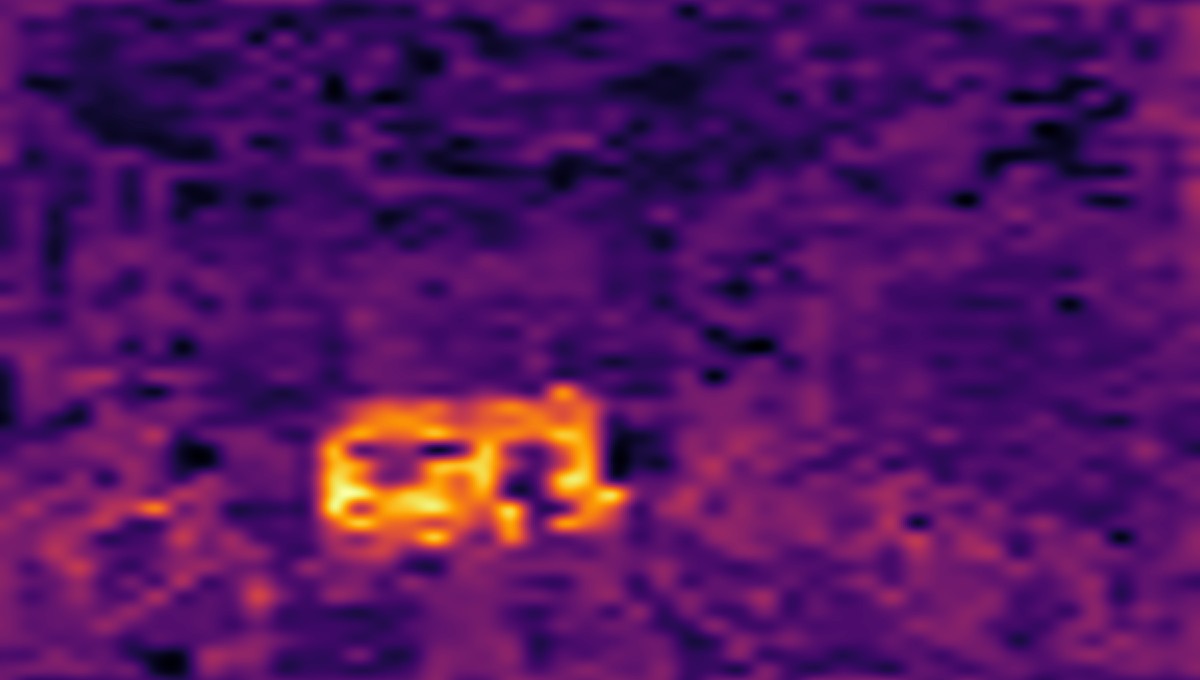}
    \subcaption[]{}
  \end{subfigure}
  \caption{Visualization of our single-pass dense feature extraction. (a) The original RGB input frame. (b) Weighted dense cosine similarity heatmap weighted for the open-vocabulary query \texttt{"an image of a chair"}. The extracted patch features provide precise semantic grounding without requiring image crops.}
  \label{fig:feature_similarity}
\end{figure}

To facilitate open-vocabulary queries, we anchor CLIP \cite{radford2021learning} embeddings into the 3D instances. The standard approach of cropping images based on instance masks is computationally prohibitive, strips away critical global context~\cite{Yu_2023_CVPR}, and often leads to misclassifications due to feature \emph{bleeding} and domain-shift~\cite{liang2023open}.

Instead, we introduce a single-pass feature extraction mechanism. Inspired by MaskCLIP~\cite{dong2023maskclip}, we extract dense, patch-level features directly from the penultimate transformer layer of a standard CLIP model during a single forward pass. Because CLIP patches are relatively large (\eg $14\times14$ pixels for ViT-L/14), simply averaging patches within a 2D SAM mask often results in diluted features, as flat background surfaces might dominate the representation. 

To mitigate this and focus on structurally relevant parts, we compute a spatial \emph{distinctiveness map} $D$ instead of a naive average. For each patch feature $f_{i,j}$ and the spatial mean of all patches $\bar{f}$, we compute the normalized residual norm as 
\begin{equation}
    D_{i,j} = \frac{\| f_{i,j} - \bar{f} \|_2}{\frac{1}{HW} \sum_{i,j} \| f_{i,j} - \bar{f} \|_2 +\epsilon}
\end{equation}
This formulation naturally assigns higher weights to patches containing unique, high-frequency information (\eg texture-rich parts) while down-weighting homogeneous backgrounds (see Figure \ref{fig:feature_similarity}). The final instance feature is obtained by aggregating the patch features within the segmentation mask, weighted by $D$.

During active refinement across frames, it is crucial that the semantic representation is not degraded by poor viewpoints. We implement an incremental view-quality fusion mechanism that assesses each observation using a combined quality score $Q$:
\begin{equation}
    Q = S_{\text{geo}} \cdot S_{\text{sem}} \cdot S_{\text{dist}}
\end{equation}
The geometric component $S_{\text{geo}} = S_{\text{size}} \cdot S_{\text{angle}}$ determines the physical observation quality. $S_{\text{size}} = \min( \lambda |M| / (H_{\text{img}} W_{\text{img}}), 1.0 )$ favors objects that occupy a significant portion of the frame area $|M|$, with $\lambda=3.3$ establishing a plateau to avoid overvaluing \emph{camera-at-the-wall} scenarios. Furthermore, 
\begin{equation}
    S_{\text{angle}} = \frac{1}{|V|} \sum_{v \in V} \max(0, -\vec{r}_v \cdot \vec{n}_v)
\end{equation}
ensures that the 3D voxel normals $\vec{n}_v$ of the detection's voxel set $V$ face the camera viewing direction $\vec{r}_v$.

The semantic score $S_{\text{sem}}$ acts as a contextual gate, filtering out segments that contradict the global scene context by calculating the clamped ReLU-activated cosine similarity between local patch features and the global image embedding. Finally, $S_{\text{dist}} = 0.5 + 0.5 \cdot \bar{D}_{\text{mask}}$ serves as a confidence multiplier derived from the average structural distinctiveness of the mask region $\bar{D}_{\text{mask}}$.

When two instances are fused, a node's feature is replaced with an observed feature if it possesses a higher quality $Q$. This ensures that each object instance is always represented by the best available local observation while being protected from dilution caused by erroneous merges \cite{kassab2024bare}.

\section{EVALUATION}

\begin{table}[t]
  \scriptsize
  \caption{
    3D open-set semantic segmentation benchmark results on subsets of the Replica~\cite{straub2019replica} and ScanNet~\cite{dai2017scannet} datasets.
  }
  \label{tab:semantic_seg}
  \centering
  \begin{tabularx}{0.45\textwidth}{lc|XXXXXX}
    \toprule 
    \multirow{2}{*}{} &
      \multicolumn{1}{c}{} &
      \multicolumn{3}{c}{Replica} &
      \multicolumn{3}{c}{ScanNet} \\
    & Methods & mAcc & mIoU & fmIoU
    & mAcc & mIoU & fmIoU \\
    \midrule
    \textit{\rotatebox[origin=c]{90}{\textit{Privileged}}} 
  & OpenFusion~\cite{yamazaki2024open}  & 0.41 & 0.30 & 0.58 & 0.67 & 0.53 & 0.64 \\
    \midrule
    \multirow{6}{*}{\rotatebox[origin=c]{90}{\textit{Zero-Shot}}}
  & ConceptFusion~\cite{jatavallabhula2023conceptfusion}  & 0.29 & 0.11 & 0.14 & 0.49 & 0.26 & 0.31\\
  & OpenMask3D ~\cite{takmaz2023openmask3d}               & - & - & - & 0.34 & 0.18 & 0.20\\
  & ConceptGraphs~\cite{Gu2024-bq}                        & 0.36 & 0.18 & 0.15 & 0.52 & 0.26 & 0.29\\
  & BBQ~\cite{linok2025beyond}                            & 0.38 & 0.27 & 0.48 & 0.56 & 0.34 & 0.36\\
  & CORE-3D~\cite{mirzaei2025core}                        & 0.38 & 0.29 & 0.56 & 0.61 & 0.36 & 0.46\\
  & DISC (Ours)    & \textbf{0.47} & \textbf{0.29} & \textbf{0.54} & \textbf{0.71} & \textbf{0.43} & \textbf{0.49} \\
  \bottomrule
  \end{tabularx}
\end{table}

To rigorously evaluate the capabilities of our DISC mapping pipeline, we conduct experiments across multiple dimensions. We benchmark our approach against state-of-the-art methods on established datasets (Replica and ScanNet) for dense segmentation, and on a subset of Habitat-Matterport 3D (\acrshort{hm3d}) for instance retrieval. Furthermore, to assess real-time scalability and robustness in large multi-story environments, we introduce a new large-scale continuous trajectory dataset based on \acrshort{hm3d}.

\subsection{Experimental Setup and Metrics} 

All \glspl{vlfm} evaluated in our study are instantiated using the OpenCLIP framework \cite{ilharco2021openclip} to ensure reproducibility and leverage large-scale open-source pre-training. Specifically, we utilize the following architectures and checkpoints:
\begin{itemize}[noitemsep, topsep=2pt, leftmargin=*]
    \item \textbf{ViT-\{B/16, L/14, H/14\}}: LAION-2B~\cite{schuhmann2022laion} pre-trained checkpoints (\textit{laion2b\_s34b\_b88k}, \textit{laion2b\_s32b\_b82k}, and \textit{laion2b\_s32b\_b79k}, respectively).
    \item \textbf{ConvNeXt-Large} (\textit{convnext\_large\_d\_320}): Utilizing the \textit{laion2b\_s29b\_b131k\_ft\_soup} weights.
    \item \textbf{EVA02-L/14} (\textit{EVA02-L-14-336}): Initialized with the \textit{merged2b\_s6b\_b61k} checkpoint.
\end{itemize}
Unless otherwise specified, all default mapping experiments and our pipeline operate using the ViT-L/14 model.
For the DINOv2 model used for segment tracking, we utilize the \textit{dinov2\_vits14\_reg} snapshot.

We evaluate our system along two primary performance axes across our experiments:
\begin{enumerate}[noitemsep, topsep=2pt, leftmargin=*]
    \item \textbf{Dense Semantic Segmentation:} To measure strict geometric classification quality, we compute the mean Accuracy (mAcc), mean Intersection over Union (mIoU), and frequency-weighted mIoU (fmIoU) metrics following standard open-vocabulary segmentation benchmarks.
    \item \textbf{Open-Vocabulary Retrieval:} To assess object-level semantic ranking, we compute the accuracy of the correct ground-truth semantic class appearing within the top-$k$ cosine-similarity predictions (Acc@$K$) and the Area Under the Curve ($AUC^{\text{top}}_{k}$) for these top-$k$ ranks, following the protocol established in \cite{werby2024hierarchical}.
\end{enumerate}

\subsection{3D Open-Set Semantic Segmentation}
\label{subsec:sem_seg_eval}

We first quantitatively evaluate the zero-shot semantic mapping capabilities of DISC on the Replica~\cite{straub2019replica} and ScanNet~\cite{dai2017scannet} datasets. We follow the benchmark methodology in \cite{linok2025beyond} to ensure a fair comparison. For Replica, we evaluate on 8 standard scenes (\textit{room0}, \textit{room1}, \textit{room2}, \textit{office0}, \textit{office1}, \textit{office2}, \textit{office3} and \textit{office4}), and similarly on 8 scenes for ScanNet (\textit{0011\_00}, \textit{0030\_00}, \textit{0046\_00}, \textit{0086\_00}, \textit{0222\_00}, \textit{0378\_00}, \textit{0389\_00} and \textit{0435\_00}). By associating each 3D point in the ground truth to its closest CLIP feature vector in our mapped scene, we obtain a point cloud with dense class labels that is directly compared against the ground truth annotations.

The quantitative results are presented in Table \ref{tab:semantic_seg}. Our mapping pipeline surpasses all other evaluated zero-shot approaches and even approaches or exceeds the privileged OpenFusion~\cite{yamazaki2024open} method, which uses the supervised SEEM Model~\cite{zou2023segment} to extract dense feature masks directly. This demonstrates that our lightweight zero-shot pipeline achieves a mask-aligned feature quality comparable to specialized, supervised models. Specifically, the high mAcc achieved by DISC proves that our dense patch-feature extraction strategy effectively solves the noise problems inherent in crop-based procedures. By extracting patch-level features directly from the vision transformer, our system obtains object-local dense feature vectors that retain sufficient global context to remain aligned with the scene without background interference or domain-shift artifacts.

\subsection{Object Level Semantics on HM3DSEM}

\begin{table}[t]
  \centering
  \resizebox{\columnwidth}{!}{%
  \begin{tabular}{@{}l|cccccc|c@{}}
    \toprule
    Method & Acc@5 & Acc@10 & Acc@25 & Acc@100 & Acc@250 & Acc@500 & AUC$^{\text{top}}_{k}$ \\
    \midrule
    VLMaps~\cite{huang2023visual}       &  0.05 &  0.17 &  0.54 & 15.32 & 26.01 & 40.02 & 56.20 \\
    ConceptGraphs~\cite{Gu2024-bq}      & 18.11 & 24.01 & 33.00 & 55.17 & 70.85 & 81.55 & 84.07 \\
    HOV-SG~\cite{werby2024hierarchical} & 18.43 & 25.73 & 36.41 & 56.46 & 69.95 & 80.86 & 84.88 \\
    \midrule
    DISC (Ours)    & \textbf{22.22} & \textbf{33.76} & \textbf{45.18} & \textbf{59.85} & \textbf{71.76} & \textbf{81.86} & \textbf{85.15} \\
    \bottomrule
  \end{tabular}%
  }
  \caption{$AUC_{top-k}$ results on the subset of HM3DSEM used by HOV-SG~\cite{werby2024hierarchical}. HOV-SG states to only consider hungarian matches with $IoU>0.5$, which we found differs in their implementation, which evaluates on all pairs after hungarian matching. }
  \label{tab:auc_hov_sg}
\end{table}

While Replica and ScanNet provide excellent dense metrics, they are predominantly limited to small, single-room scenes. To evaluate object-level retrieval, we leverage the \acrshort{hm3d} dataset \cite{Yadav2023CVPR}. We first benchmark our approach on the identical 10-scene subset and trajectories used by HOV-SG~\cite{werby2024hierarchical}. We evaluate the open-vocabulary query performance using our established retrieval metrics (Acc@$k$ and $AUC^{\text{top}}_{k}$). 

The authors of HOV-SG state that they only consider predicted objects with an $IoU>0.5$ against the ground truth. However, their official implementation evaluates all pairs after Hungarian matching without this strict threshold. For a fair comparison, we adopted this exact matching protocol. As shown in Table \ref{tab:auc_hov_sg}, DISC outperforms HOV-SG and ConceptGraphs across all values of $k$, achieving a slight improvement in overall $AUC^{\text{top}}_{k}$ ($0.27\%$). Notably, we obtain substantial improvements of $3.79\%$ and $13.63\%$ in the strict Acc@$5$ and Acc@$10$ metrics, which are the most relevant bounds for practical downstream robotic tasks.

\subsection{Large-Scale Dataset Generation}
\label{subsec:dataset_generation}

\begin{figure}[t]
  \centering
  \includegraphics[width=\linewidth]{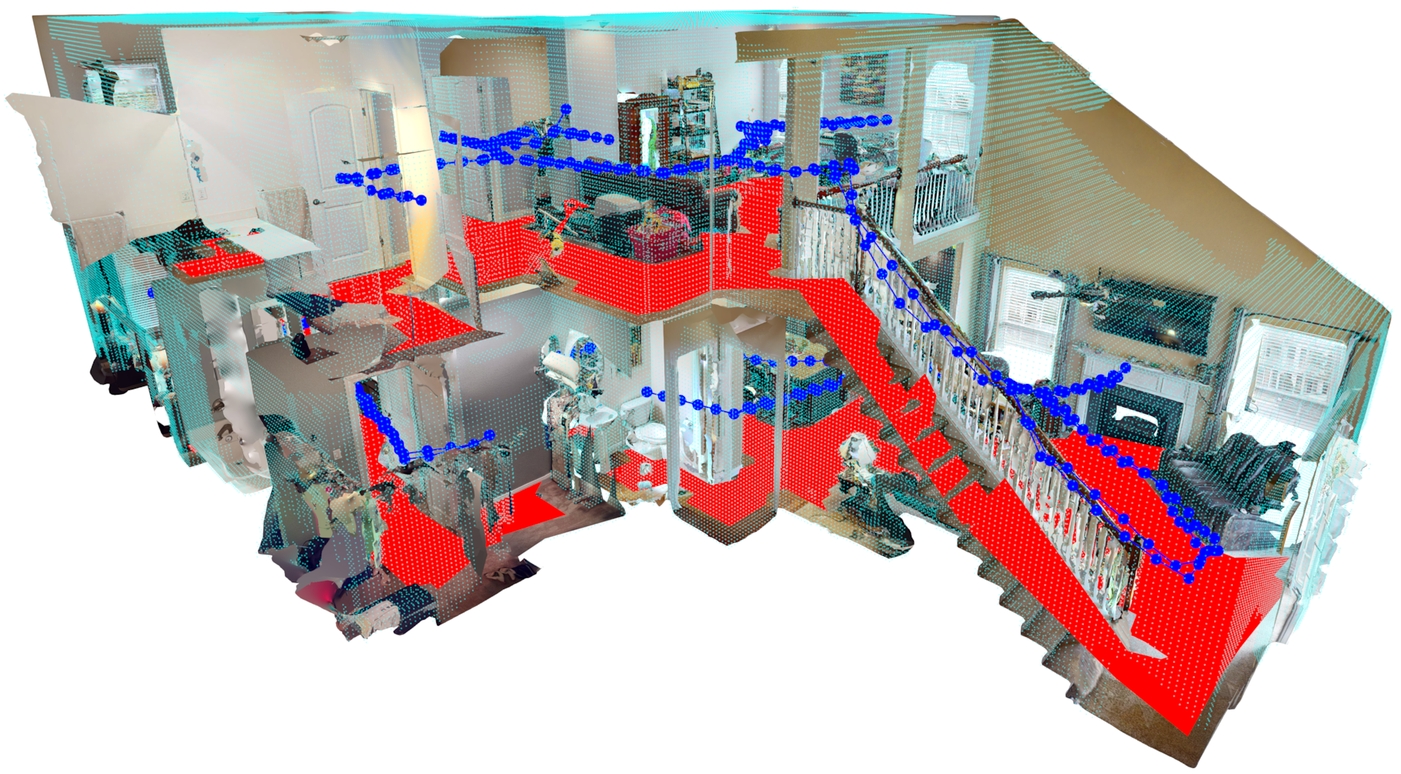}
  \caption{Generated trajectory from \acrshort{hm3d} scene \textit{00800}. Red: navigation mesh. Blue lines: generated trajectory. Teal points: coverage analysis from ray tracing as down sampled voxel grid.}
  \label{fig:dataset_poses}
\end{figure}

HOV-SG uses  manually recorded random walks for 10 out of 36 scenes in the validation split of HM3DSEM. 
To the best of our knowledge, there is no publicly available set of trajectories that covers all scenes, the train and validation split of HM3DSEM.
The underlying Habitat simulator~\cite{savva2019habitat} automatically computes navigation meshes for an agent's configuration space, given a 3D scene. This serves as a realistic proxy for a well-conditioned continuous robotic exploration, which we utilize to generate continuous RGB-D trajectories that naturally cover most of the simulated environments. 

We first extract the largest connected component of the navigation mesh, apply a Voronoi decomposition~\cite{Hughes2022-fg, werby2024hierarchical} and the brushfire algorithm to extract local basins (places).
By computing the geodesic distance via A* between adjacent places on the medial axis, we build a graph. The exploration trajectory is then efficiently derived by solving the Chinese Postman Problem, guaranteeing optimal edge traversal (see Figure~\ref{fig:dataset_poses}).
Finally, Habitat's builtin pathfinder and greedy follower generate smooth trajectories via B\'ezier curves.

\subsubsection{Dataset: Configuration and Evaluation}
\begin{figure}[b]
  \centering
  \begin{subfigure}[c]{0.49\linewidth}
    \centering
    \includegraphics[width=\linewidth]{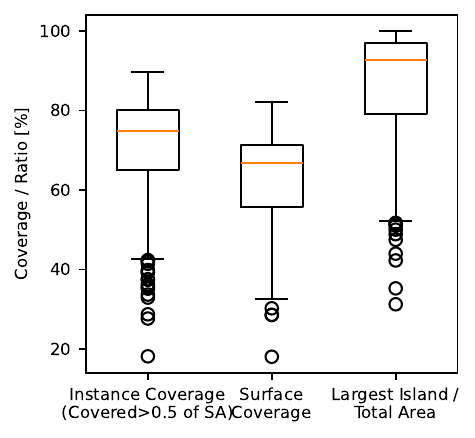}
    \caption{\label{fig:visibility}}
  \end{subfigure}
  \begin{subfigure}[c]{0.49\linewidth}
    \centering
    \includegraphics[width=\linewidth]{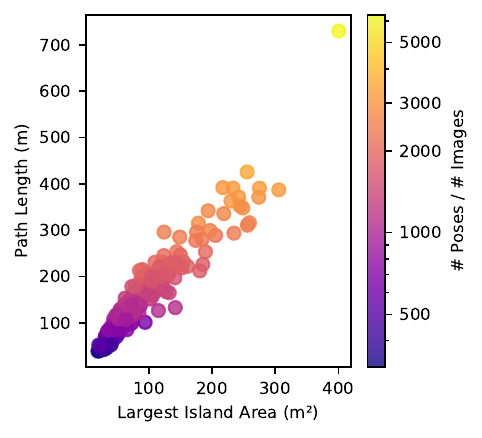}
    \caption{\label{fig:tour_length}}
  \end{subfigure}
  \caption{(a) Ratio of covered objects ($Area>50\%$ of surface covered) and ratio of covered surface for our generated trajectories for \acrshort{hm3d}. High variance is explained by the fact that the largest connected component of the navigation mesh from Habitat only covers the scene partially (Largest Island/Total Area).
  (b) Tour length plotted in relation to the navigable area, with color encoding the resulting dataset size (= simulation steps required).}
\end{figure}

For dataset generation, we use an agent with a sensor height of \SI{1.2}{\metre} and a discrete action space consisting of \textit{move\_forward} \SI{0.25}{\metre} and \textit{turn\_left/turn\_right} \SI{15}{\degree} to sample discrete keyframes along a simple exploration strategy where the agent looks straight ahead.
We use Habitat's standard configuration for navigation mesh generation and apply the approach detailed in sec. \ref{subsec:dataset_generation} to generate trajectories for the train (145) and val (36) splits of \acrshort{hm3d}.

To evaluate the quality of generated datasets and assess how well a generated trajectory covers the underlying scene we evaluate the coverage of our resulting trajectories.
We estimate a local voxel grid for all poses from the full resolution semantic meshes using ray tracing~\cite{mock2023rmagine} and compute the per instance and surface coverage by comparing these local grids against the full voxel map (See Table~\ref{tab:instance_visibility}). 

This helps to identify objects that can never be observed along the trajectory and establish a lower bounding on the overall visibility of each instance. As Figure~\ref{fig:visibility} shows, the resulting coverage varies significantly over the whole dataset.
We envision several potential uses: as a measure of quality for trajectories, for determining fair ceilings for agent or system performance, and/or for evaluating the quality of the underlying scenes themselves.

\begin{table}[h]
\centering
\resizebox{\columnwidth}{!}{%
\begin{tabular}{c l c r r r}
\hline
\textbf{ID} & \textbf{Category} & \textbf{Region} & \textbf{Model Voxels} & \textbf{Covered Voxels} & \textbf{Coverage (\%)} \\
\hline
113 & chair & 2 & 9773 & 8340 & 85.34 \\
114 & table & 2 & 23818 & 22922 & 96.24 \\
115 & flower vase & 2 & 2656 & 2599 & 97.85 \\
116 & ceiling & 3 & 317994 & 239945 & 75.46 \\
117 & wall & 3 & 22406 & 15816 & 70.59 \\
118 & wall & 3 & 10972 & 10237 & 93.30 \\
119 & door & 3 & 17056 & 17038 & 99.89 \\
120 & calendar & 3 & 1555 & 1555 & 100.00 \\
121 & kitchen shelf & 3 & 13980 & 8382 & 59.96 \\
\hline
\end{tabular}
}

\caption{An excerpt of the surface coverage analysis details how well regions and object instances have been covered for \textit{00891}. \label{tab:instance_visibility}}

\end{table}

\subsection{Mapping in Large-Scale Environments}

\begin{figure}[tb]
  \centering
  \includegraphics[width=\linewidth]{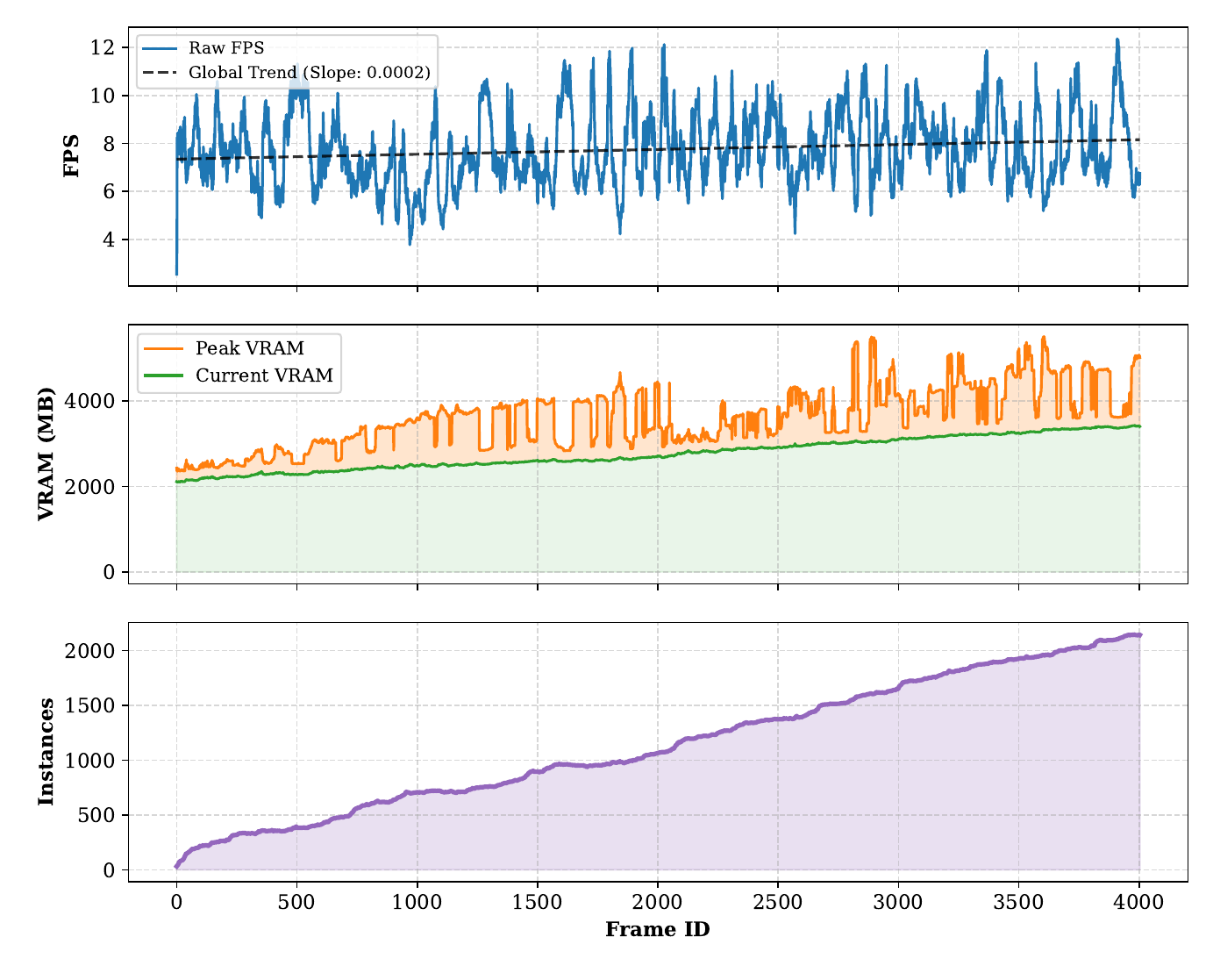}
  \caption{Performance of the mapping pipeline on the exemplary \acrshort{hm3d} scene \textit{00849}.}
  \label{fig:map_performance}
\end{figure}

To demonstrate the robustness and scalability of our approach in expansive, visually complex environments, we evaluate our pipeline on our generated \acrshort{hm3d} trajectories. Unlike single-room datasets, these multi-room and multi-story facilities test a system's ability to maintain semantic consistency over long horizons without succumbing to feature dilution or catastrophic forgetting if objects are re-observed from very different view poses. 

Table \ref{tab:hm3dsem_evaluation} reports the dense zero-shot segmentation and open-vocabulary retrieval metrics across the \textit{val} and \textit{train} splits. Since these specific dense mapping trajectories establish a novel benchmark setup, we report the absolute metrics to provide a comprehensive baseline for future research. The results demonstrate good semantic stability at scale: across the combined splits, our approach achieves an mAcc of $0.33$ and an fmIoU of $0.42$, consistent with the smaller validation subset (mAcc $0.34$, fmIoU $0.45$). Similarly, the retrieval performance remains robust, maintaining an $AUC^{\text{top}}_{k}$ of $0.84$ across all scenes. This confirms that our view-quality gating ($Q$) successfully protects the semantic representations from deteriorating, ensuring that aggregated object features remain sharp even in very large environments where the same object instance is observed many times.
Beyond semantic accuracy, the extensive nature of the \acrshort{hm3d} dataset also highlights the system's operational efficiency. To evaluate the run-time behavior of our approach, we mapped a 4000-frame trajectory of a single scene on a desktop system equipped with an Intel Core i7-13700K CPU, 64\,GB of RAM, and an NVIDIA RTX 5080 GPU (see Figure~\ref{fig:map_performance}). 

As illustrated, the number of tracked 3D instances grows continuously as the camera explores new areas of the scene. Despite this accumulating map complexity, our pipeline maintains a consistent frame processing rate (top plot). While the immediate FPS fluctuates naturally depending on complexity of the current view (\eg the number of visible segments), on larger scales it remains nearly constant. This confirms that the computational cost of our active voxel-based refinement and single-pass feature extraction remains strictly bounded and do not degrade over time.
Furthermore, the GPU memory profiling (middle plot) demonstrates an efficient and predictable VRAM footprint as well.

Unlike conventional systems that require the mapping process to pause for heavy offline refinement as the number of instances in the map grows large, our active voxel-based update mechanism seamlessly scales to thousands of instances. This verifies that our architecture is not only semantically robust but highly viable for continuous, real-time operation on mobile robots exploring large-scale facilities.

\begin{table}[t]
  \centering
  \resizebox{\columnwidth}{!}{%
  \begin{tabular}{l|cccccc@{}}
    \toprule
    Split & mAcc & mIoU & fmIoU &  Acc@5 & Acc@10 & AUC$^{\text{top}}_{k}$ \\
    \midrule
    val & 0.34 & 0.20 & 0.45 & 21.45 & 31.63 & 85.93 \\
    train & 0.33 & 0.17 & 0.42 & 18.87 & 28.61 & 84.10 \\
    combined & 0.33 & 0.18 & 0.42 & 19.09 & 29.64 & 84.41 \\
    \bottomrule
  \end{tabular}%
  }
  \caption{Comparison of dense zero-shot segmentation and open-vocabulary retrieval metrics across the large scale \acrshort{hm3d} dataset.}
  \label{tab:hm3dsem_evaluation}
\end{table}

\subsection{Comparison of Vision-Language Backbones}

\begin{table}[t]
  \centering
  \resizebox{\columnwidth}{!}{%
  \begin{tabular}{lc|ccccc@{}}
    \toprule
    & Method & mAcc & mIoU & fmIoU &  Acc@5 & Acc@10 \\
    \midrule
    \multirow{5}{*}{\rotatebox[origin=c]{90}{\textit{patch features}}}    
    & ViT-B/16 & 0.30 & 0.20 & 0.41 & \textbf{25.78} & \textbf{35.13} \\
    & ViT-L/14 & \textbf{0.34} & \textbf{0.20} & \textbf{0.45} & \textbf{21.96} & \textbf{33.87} \\
    & ViT-H/14 & 0.34 & 0.22 & 0.39 & 16.32 & 24.64 \\
    & ConvNeXt-L & 0.30 & 0.33 & 0.17 & 13.06 & 18.93 \\
    & EVA02-L/14 & 0.25 & 0.15 & 0.33 & 13.60 & 20.62 \\
    \midrule
    \multirow{5}{*}{\rotatebox[origin=c]{90}{\textit{crop features}}}    
    & ViT-B/16 & 0.28 & 0.15 & 0.36 & 16.11 & 23.00 \\
    & ViT-L/14 & 0.34 & 0.17 & 0.39 & 16.84 & 25.15 \\
    & ViT-H/14 & 0.31 & 0.17 & 0.38 & 16.46 & 24.17 \\
    & ConvNeXt-L & 0.38 & 0.18 & 0.40 & 15.96 & 24.66 \\
    & EVA02-L/14 & 0.41 & 0.23 & 0.46 & 20.25 & 29.37 \\
    \bottomrule
  \end{tabular}%
  }
  \caption{Comparison of dense zero-shot segmentation and open-vocabulary retrieval metrics across various CLIP backbone architectures using crop-based and single-pass patch feature extraction.}
  \label{tab:comparison_backbones}
\end{table}

To identify the optimal foundation model and validate our single-pass feature integration, we conduct a comparative analysis of different CLIP backbone architectures on our \acrshort{hm3d}-based dataset (val split). We adapt each model to follow our single-pass patch feature extraction approach and compare them against the classical crop-based baseline (Table \ref{tab:comparison_backbones}). We report dense 3D segmentation metrics (mAcc, mIoU, fmIoU) alongside open-vocabulary retrieval metrics (Acc@5, Acc@10), measuring whether the correct semantic class for matched objects ranks within the top-$k$ predictions.

Our results yield three key insights:

\textbf{(1)} For standard Vision Transformers (ViT-B, ViT-L, ViT-H), our single-pass patch extraction consistently outperforms or matches the crop-based approach in dense geometric metrics. For instance, using ViT-L/14, patch extraction increases the fmIoU from 0.39 to 0.45 while maintaining the same mAcc of 0.34. This demonstrates that avoiding artificial crops prevents domain-shift artifacts and feature bleeding, causing the features to align much better with the physical boundaries of the objects. Conversely, models that rely upon final global average pooling to align visual features with the text space—such as the CNN-based ConvNeXt or EVA02 \cite{fang2024eva}—struggle significantly with intermediate patch extraction. Extracting spatial maps from ConvNeXt-L drastically drops its semantic retrieval (Acc@5 falls from 0.16 to 0.13) and dense classification capabilities (fmIoU drops from 0.40 to 0.17). This highlights that our zero-shot approach works best with standard contrastive ViT architectures, where intermediate tokens maintain stronger semantic text-alignment.

\textbf{(2)} The shallower ViT-B/16 model achieves the highest soft-retrieval scores (Acc@5 of 0.26) but gets outperformed by the deeper ViT-L/14 model on dense tasks (mAcc 0.34 vs. 0.30, and fmIoU 0.45 vs. 0.41). We hypothesize that due to its fewer transformer layers, ViT-B/16 retains less spatially smoothed, more localized signatures that are advantageous for instance retrieval. However, it appears to lack the deeper semantic discriminative capacity of larger models, which becomes crucial for strict dense classification.

\textbf{(3)} Consequently, we selected ViT-L/14 as the default foundation model for the DISC pipeline. In our evaluations, it offered the highest dense semantic performance (fmIoU 0.45) while maintaining highly competitive open-vocabulary retrieval capabilities. Additionally, it is a robust, well-tested architecture proven to work reliably with the MaskCLIP-style patch extraction, in contrast to alternative architectures where bypassing the global pooling layer might result in unstable semantic representations.

\section{CONCLUSIONS}

In this paper, we presented DISC (Dense Integrated Semantic Context), a fully GPU-accelerated mapping architecture designed to bridge the scalability gap in open-set 3D semantic mapping. By transitioning from fast but imprecise bounding-box heuristics to dense, voxel-based overlap calculations, our system eliminates the need for computationally prohibitive offline refinement stages, robustly fusing instances on-the-fly during exploration. Furthermore, we addressed a significant bottleneck in Vision-Language feature extraction by integrating a single-pass, distance-weighted feature extraction approach. Guided by DINOv2 tracking and a geometric view-quality scoring mechanism, DISC anchors high-fidelity, open-vocabulary embeddings directly into the 3D map without the context loss and latency associated with traditional image cropping. We demonstrated the efficacy and scalability of our approach on standard benchmarks and a novel, large-scale HM3D-based dataset tailored for continuous multi-story mapping.

\subsection{Limitations} 

Despite its strong performance, our approach has certain limitations. First, the pipeline fundamentally relies on the quality of the underlying 2D class-agnostic segmentation model (\eg FastSAM). Consistent under-segmentation or failure to detect highly occluded objects in 2D naturally propagates to the 3D map. Second, the spatial resolution of the semantic embeddings is bounded by the patch size of the underlying Vision Transformer (\eg $14 \times 14$ pixels for ViT-L/14). While our distinctiveness map $D$ effectively mitigates background bleeding, extracting pure semantic representations for extremely small or thin structures (\eg cables or thin poles) remains challenging from distant viewpoints. Finally, the current voxel-integration mechanism implicitly assumes a predominantly static environment.

\subsection{Future Work} 

In future work, we aim to extend the DISC architecture by shifting to active, language-driven robotic exploration as outlined in [\emph{Ref. suppressed for anonymity}]. By utilizing the fast, incremental nature of DISC, we aim for robots to actively search for open-vocabulary targets in massive environments, deciding on-the-fly which unexplored areas are semantically most promising and automatically select appropriate exploration poses for any object.
Other promising directions are the utilization of the dense semantic features for downstream tasks like localization, embodied question answering and planning \& acting.

\section*{ACKNOWLEDGMENTS}
\blackout{%
This work is supported by the \censor{ExPrIS and LIEREx projects} through grants from the German Federal Ministry of Research, Technology and Space (BMFTR) with grant numbers 16IW23001 and 16IW24004.
The DFKI Niedersachsen (DFKI NI) is sponsored by the Ministry of Science and Culture of Lower Saxony and the Volkswagen Foundation.
} 

\bibliographystyle{IEEEtran}
\bibliography{bibliography}

\end{document}